\documentclass[letterpaper]{article} 
\usepackage[submission]{aaai25}  
\usepackage{times}  
\usepackage{helvet}  
\usepackage{courier}  
\usepackage[hyphens]{url}  
\usepackage{graphicx} 
\usepackage{natbib}  
\usepackage{caption} 
\usepackage{algorithm}
\usepackage{algorithmic}
\usepackage{amssymb}
\usepackage{amsmath}
\usepackage{booktabs}

\usepackage{newfloat}
\usepackage{listings}
\DeclareCaptionStyle{ruled}{labelfont=normalfont,labelsep=colon,strut=off} 
\floatstyle{ruled}
\newfloat{listing}{tb}{lst}{}
\floatname{listing}{Listing}
\title{VGG-Tex: A Vivid Geometry-Guided Facial Texture Estimation Model for High  Fidelity Monocular 3D Face Reconstruction}
\author{
   Haoyu Wu\textsuperscript{\rm 1},
   Ziqiao Peng\textsuperscript{\rm 1},
   Yunfei Cheng\textsuperscript{\rm 1},
   Xukun Zhou\textsuperscript{\rm 1},
   Jun He\textsuperscript{\rm 1},
   Hongyan Liu\textsuperscript{\rm 2},
   Zhaoxin Fan\textsuperscript{\rm 3,4*}
}
\affiliations{
   \textsuperscript{\rm 1}Renmin University of China\\
   \textsuperscript{\rm 2}Tsinghua University\\
   \textsuperscript{\rm 3}Beijing Advanced Innovation Center for Future Blockchain and Privacy Computing, Institute of Artificial Intelligence, Beihang University\\
   \textsuperscript{\rm 4}Beijing Academy of Blockchain and Edge Computing \\
   \{wuhaoyu556, ziqiaopeng, chengyunfei, hejun\}@ruc.edu.cn, liuhy@sem.tsinghua.edu.cn, zhaoxinf@buaa.edu.cn
}

\usepackage{bibentry}

\begin{document}

\maketitle
\footnotetext{*Corresponding authors}
\begin{abstract}
3D face reconstruction from monocular images has promoted the development of various applications such as augmented reality. Though existing methods have made remarkable progress, most of them emphasize geometric reconstruction, while overlooking the importance of texture prediction. To address this issue, we propose VGG-Tex, a novel Vivid Geometry-Guided Facial Texture Estimation model designed for High Fidelity Monocular 3D Face Reconstruction. The core of this approach is leveraging 3D parametric priors to enhance the outcomes of 2D UV texture estimation. Specifically, VGG-Tex includes a Facial Attributes Encoding Module, a Geometry-Guided Texture Generator, and a Visibility-Enhanced Texture Completion Module. These components are responsible for extracting parametric priors, generating initial textures, and refining texture details, respectively. Based on the geometry-texture complementarity principle, VGG-Tex also introduces a Texture-guided Geometry Refinement Module to further balance the overall fidelity of the reconstructed 3D faces, along with corresponding losses. Comprehensive experiments demonstrate that our method significantly improves texture reconstruction performance compared to existing state-of-the-art methods.
\end{abstract}

%

\section{Introduction}\label{sec:intro}
3D face reconstruction stands as a pivotal challenge within the field of computer vision, endeavoring to 3D depictions of faces from mere monocular 2D  images. This endeavor finds its utility in a myriad of downstream applications, from enhancing speech-driven facial animations \cite{peng2023emotalk,selftalk,peng2023synctalk} to enriching the immersive realms of 3D video games\cite{wang2006face,lin2021meingame} and augmenting the interactivity in augmented \cite{wei2022assessing,fan2022object}  and virtual reality \cite{,fan2024everything2motion,thies2016facevr} experiences.

Over the past decades, numerous studies \cite{deep3d,DECA,MICA,HRN,hiface,denselandmark} have been introduced. For instance, DECA \cite{DECA} stands out as a significant work utilizing unlabeled face images for unsupervised 3D face reconstruction, while MICA \cite{MICA} estimates human face shapes from a single image using a supervised approach that combines various 2D, 2D/3D, and 3D datasets. Although these methods have shown impressive results, they primarily focus on geometric reconstruction.

\begin{figure}
    \centering
    \includegraphics[width=0.5\textwidth]{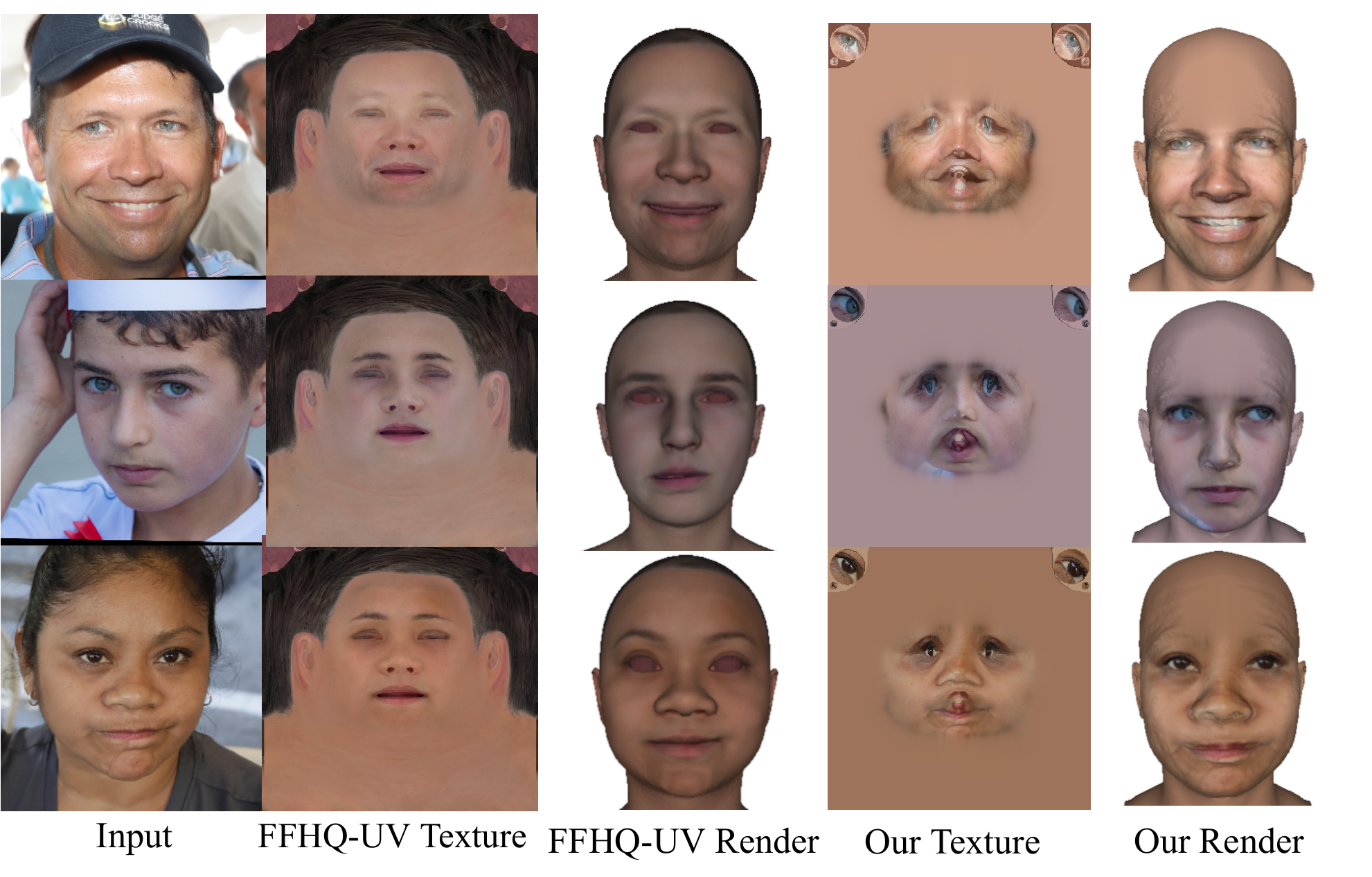}
    \captionof{figure}{\textbf{Intuition of VGG-TEX.} A comparison between FFHQ-UV and our method demonstrates a fact that the texture of a 3D face can greatly affect how humans perceive it, even if the geometric details are not very fine.}
    \label{head}
    \vspace{-0.2in}
\end{figure}%

However, it is commonly understood, as depicted in Fig. \ref{head}, that the texture of a 3D face can greatly affect how humans perceive it. This means that even if the geometric details are not very fine, having better textures can still greatly improve the visual experience. As a result, recent research \cite{FairAlbedo,AlbedoGAN,uvgan,ganfit,deep3d,ffhquv,ostec} has started to look into improving texture estimation quality. Yet, these methods often rely on annotated UV texture datasets to train image generators or use optimization-based approaches to create detailed UV mappings. This leads to high costs in gathering datasets with UV maps and significant resource use in optimization processes. Therefore, finding an efficient and effective way to estimate high-quality texture maps for high fidelity monocular 3D faces is still an open question.

Inspired by the discussion above, this paper aims to simultaneously reconstruct high-quality geometry and texture to facilitate 3D face reconstruction. Since the task of facial geometry reconstruction has been extensively studied, as mentioned earlier, this paper primarily focuses on improving facial texture estimation performance. To achieve this, we propose VGG-Tex, a novel model designed for high fidelity monocular 3D face reconstruction, which guides texture estimation using detailed geometric informations.

Distinct from existing methods that solely utilize direct information from images for human facial texture estimation, VGG-Tex incorporates 3D parametric priors to enhance the results of 2D UV texture estimation. Specifically, VGG-Tex introduces three key components: a Facial Attributes Encoding Module, a Geometry-Guided Texture Generator, a the Visibility-Enhanced Texture Completion Module.
The Facial Attributes Encoding Module and the Geometry-Guided Texture Generator form the dual-branch network architecture of VGG-Tex. Within the Facial Attributes Encoding Module, VGG-Tex predicts the parameters of the FLAME model \cite{FLAME} for geometry reconstruction, and also estimates a latent geometry embedding, which aids subsequent texture estimation. In the  Geometry-Guided Generator, VGG-Tex employs a vision Transformer \cite{ViT} encoder and a texture decoder for UV texture estimation. This process is supported by the previously mentioned latent geometry embedding, showcasing a method of geometry-guided facial texture estimation. Expanding beyond these modules, the Visibility-Enhanced Texture Completion Module incorporates random masks on input images to simulate obscured parts, thereby equipping the model with the capability to inpaint these invisible texture areas effectively. Furthermore, adhering to the geometry-texture complementarity principle \cite{oh2001image,blanz2023morphable}, VGG-Tex introduces a Texture-guided Geometry Refinement training strategy to further enhance the overall fidelity of the reconstructed 3D faces, accompanied by corresponding losses, ensuring a harmonious balance in the reconstructed outputs.

To validate the effectiveness of VGG-Tex, we undertake both qualitative and quantitative evaluations on several benchmark datasets, including FHQ-UV, VGGFace2, and NoW. Through extensive testing, our findings reveal that VGG-Tex markedly enhances texture reconstruction performance, surpassing existing state-of-the-art methods.

Our contribution can be summarized as:
\begin{itemize}
\item We introduce VGG-Tex, a method designed for high-quality geometry and texture reconstruction in the field of monocular 3D face reconstruction, employing the concept of geometry-guided texture estimation.
\item We develop three innovative modules: the Facial Attributes Encoding Module, the Geometry-Guided Texture Generator, and the Visibility-Enhanced Texture Completion Module, all aimed at achieving high-fidelity 3D facial texture estimation.
\item We also introduce the Texture-guided Geometry Refinement training strategy along with a corresponding training loss for VGG-Tex, founded on the principle of geometry-texture complementarity.
\end{itemize}

\section{Related work}
\label{sec:Related work}
In this paper, we focus primarily on high-fidelity monocular 3D face reconstruction, with a particular emphasis on human facial texture estimation. To set the stage, we first review two relevant areas of study: geometry estimation and texture estimation in monocular 3D face reconstruction.

\subsection{Geometry Estimation in Monocular 3D Face Reconstruction}
Monocular 3D Face Reconstruction is a significant yet challenging task, especially relevant in applications such as augmented reality. Early methodologies (\cite{deep3d,DECA,MICA,HRN,hiface,denselandmark}) primarily focus on enhancing the quality of geometry reconstruction. For instance, \cite{deep3d} introduces a deep learning approach for weakly supervised 3D face reconstruction, while DECA \cite{DECA} implements a cycle-loss for unsupervised parametric 3D face estimation. MICA \cite{MICA} concentrates on metrically accurate reconstruction in a supervised training context. More recently, HRN \cite{HRN} develops a Hierarchical Representation Network to achieve accurate and detailed face reconstruction from in-the-wild images. Concurrently, HiFace \cite{hiface} proposes a method to learn both static and dynamic details to improve geometry reconstruction.

Although these methods demonstrate commendable performance, as previously noted, geometry is not the sole factor influencing how humans perceive reconstructed faces. Texture is equally important. In this paper, we explore the crucial task of texture estimation, while also leveraging the results of geometry reconstruction as a guiding framework.

\subsection{Texture Estimation in Monocular 3D Face Reconstruction}

Accurately representing facial textures is a pivotal aspect of human face and head reconstruction from monocular RGB images. Most existing methods, such as those based on 3DMM \cite{TRUST, DECA}, typically deduce coefficients within a statistical, low-dimensional linear UV space \cite{BFM,FLAME,AlbedoMM}. Given that this linear UV space represents only a subset of the RGB image space, they \cite{DECA,TRUST} inherently struggle to capture high-frequency details, such as wrinkles. To tackle the issue, several methods \cite{AlbedoGAN,FairAlbedo,ganfit,uvgan,UVLDM} have embraced the robust representational capabilities of generative models \cite{stylegan,stylegan2} to refine the representation issue and produce more realistic UV maps. AlbedoGAN \cite{AlbedoGAN}, for instance, learns to generate albedo maps that correspond to the StyleGAN \cite{stylegan,stylegan2} latent space, initially trained using a small-scale texture dataset. FFHQ-UV \cite{ffhquv} introduces a technique utilizing StyleGAN-based facial image editing approaches to generate multi-view normalized face images from single-image inputs, enhancing texture estimation. FairAlbedo \cite{FairAlbedo} designs an ID2Albedo module to produce the identity albedo map of a person from the ArcFace \cite{arcface} latent space, also trained with a private texture dataset.

Although these methods have advanced the field, they either necessitate resource-intensive optimization or rely on costly manually annotated datasets. In this paper, we introduce VGG-Tex, a method designed for both efficient and effective UV texture estimation under the concept of geometry-guided facial texture prediction.

\begin{figure*}[h]
    \centering
    \captionsetup{type=figure}
    \includegraphics[width=\textwidth]{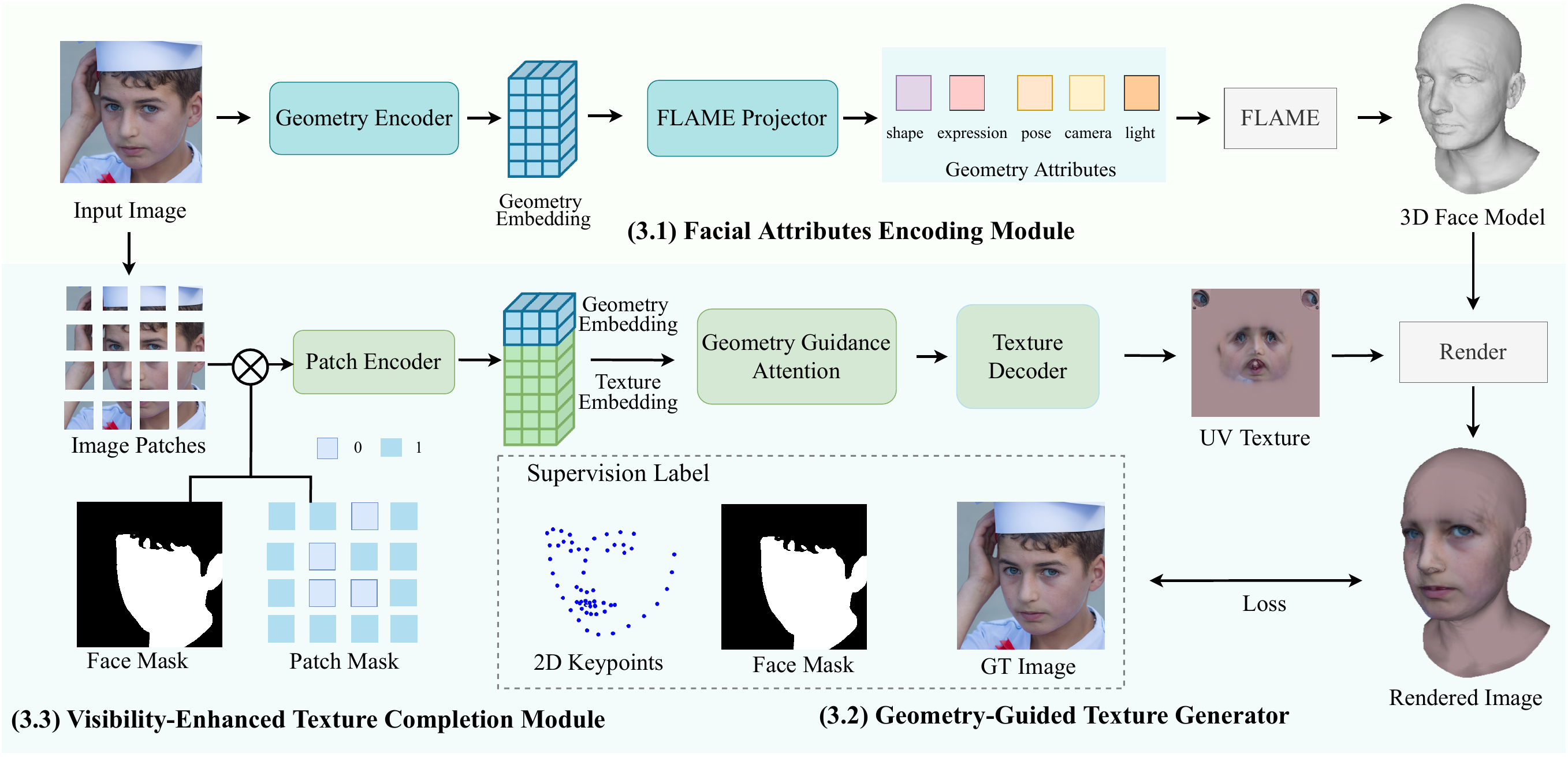}
    \captionof{figure}{\textbf{Illustration of VGG-Tex architecture.}  VGG-Tex is consisted of a dual-branch architecture. The top branch is a Facial Attributes Encoding Module for latent geometry extractiuon and 3D face geometry prediction; while the bottom branch is a Geometry-Guided Generator that takes the image and geometry guidance as input for UV texture estimation. A  During training, the Visibility-Enhanced Texture Completion Module plays a critical role by adding random masks to input images, simulating obscured parts often encountered in wild scenarios.}
    \label{fig:2 model overview}
    \vspace{-0.2in}
\end{figure*}

\section{Method}

As previously mentioned, VGG-Tex comprises a dual-branch network architecture, as illustrated in Fig. \ref{fig:2 model overview}. The first branch, known as the Facial Attributes Encoding Module, processes the original human face image. This module predicts the 3D FLAME parameters to first reconstruct the geometry of the 3D head. Simultaneously, it extracts a latent geometry embedding from the image, serving as the geometric guidance for subsequent modules. Subsequently, the lower branch, termed the Geometry-Guided Generator, utilizes the tokenized face image. It employs a vision Transformer \cite{ViT} encoder to initially extract the texture embedding. This embedding, in conjunction with the latent geometry, feeds into subsequent submodules for UV texture prediction. The resulting UV texture map, coupled with the 3D geometric model, is used to render an output image. This output then serves as a supervision signal, which is compared with the initial image, the face mask, and the 2D keypoints to train the network. During training, the Visibility-Enhanced Texture Completion Module plays a critical role by adding random masks to input images, simulating obscured parts often encountered in wild scenarios, thereby enhancing performance.

\begin{figure}
    \centering
    \captionsetup{type=figure}
    \includegraphics[width=0.5\textwidth]{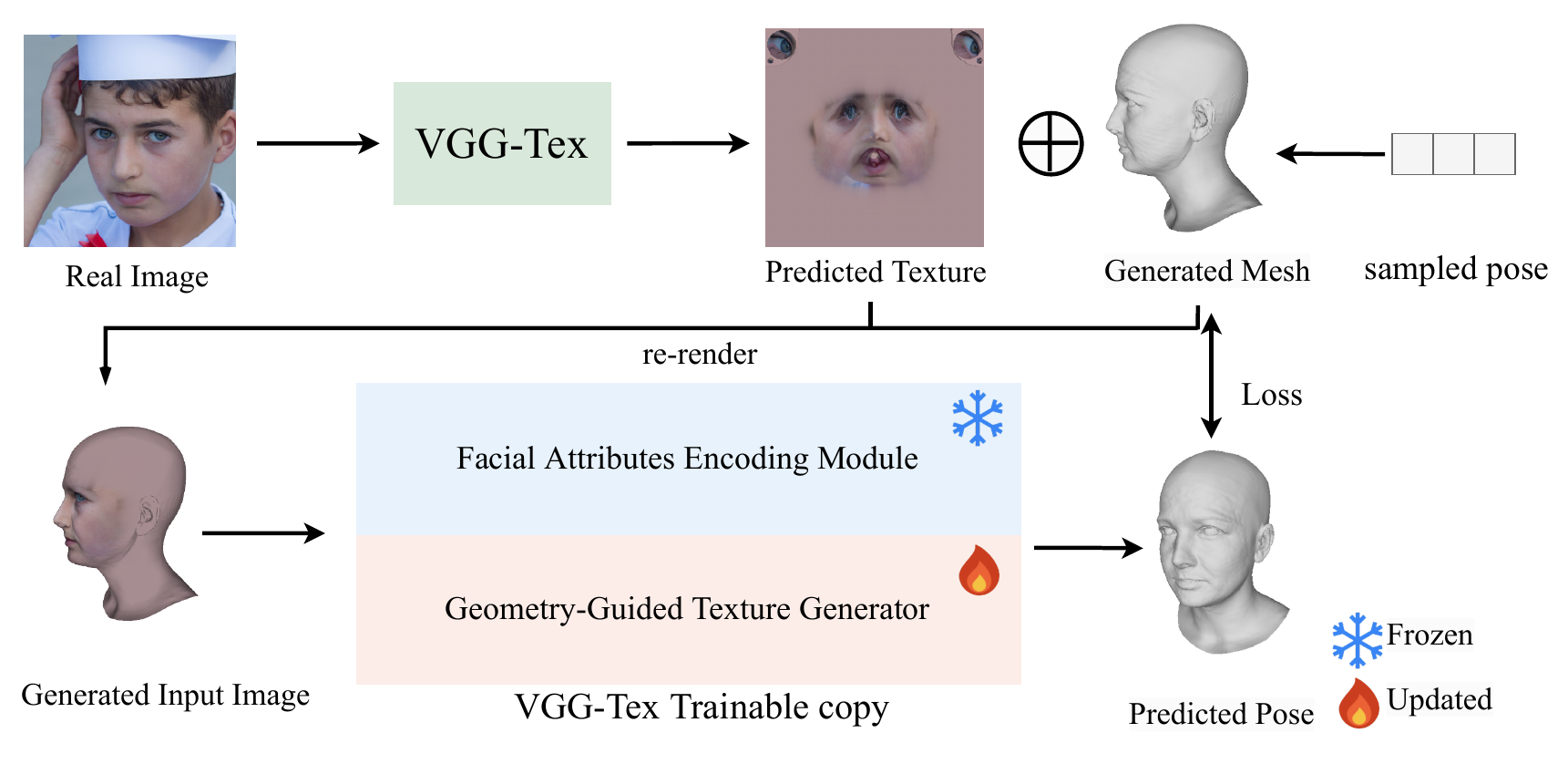}
    \captionof{figure}{\textbf{Illustration of Texture-Guided Geometry Refinement training stage.} The procedure initiates with the reconstruction of a 3D mesh and UV texture from a given input image. This is followed by sampling a head pose. The projection of the 3D head model onto the 2D image space, utilizing the sampled challenging pose, culminates in the creation of an augmented input image, denoted as \(\mathcal{I}_{r}\). This augmented image is then inputted into the geometry prediction module, which refines the pose and camera parameters by optimizing the 2D landmarks. This optimization allows the model to more effectively accommodate head pose.}
    \label{geometry2texture}
     \vspace{-0.2in}
\end{figure}

Furthermore, to elevate reconstruction quality, VGG-Tex integrates a Texture-Guided Geometry Refinement training stage, adhering to the principle of geometry-texture complementarity, as delineated in Fig. \ref{geometry2texture}.

We will now delve into the specifics of the Facial Attributes Encoding Module, the Geometry-Guided Generator, and the Visibility-Enhanced Texture Completion Module. Subsequently, we will discuss the Texture-Guided  Geometry Refinement Module and the training loss, providing a comprehensive understanding of each component.

\subsection{Facial Attributes Encoding Module}
\label{sec:FAE}

As shown in Fig. \ref{fig:2 model overview} (top), the Facial Attribute Encoder Module(FAEM) is adeptly trained to infer FLAME parameters from single-input face images. This module integrates a vision transformer network followed by a MLP as a mapping network. The resultant embedding encapsulates the shape attribute $s \in \mathbb{R}^{300}$, expression $e \in \mathbb{R}^{50}$, and pose $p \in \mathbb{R}^6$, as delineated by the following equation:

\begin{equation}
    \{s, e, p\} = FAEM(I_{\text{input}}).
\end{equation}

Following this, the FLAME model \(M \in \mathbb{R}^{5023 \times 3}\) is reconstructed from the predicted parameters \(s\) and \(e\) using the equation:
\begin{equation}
    M(s, e) = T_{\text{head}} + s B_{\mathcal{S}} +e B_{e}.
\end{equation}

In this context, \(T_{\text{head}}\) signifies the template vertices of the FLAME model, while \(B_{\mathcal{S}}\) and \(B_{e}\) represent the principal components corresponding to the shape and expression, respectively. The pose parameter is instrumental in controlling the jaw and neck pose of the human. The camera parameters, encompassing scale (1 dimension), rotation (3 dimensions), and translation (3 dimensions), are also crucial for accurate model alignment.

Simultaneously, a light encoder is employed to estimate the light condition \(L \in \mathbb{R}^{9 \times 3}\). This module captures lighting information through spherical harmonic coefficients from nine directions of RGB lights, providing a compact yet expressive representation of the lighting environment. This enables the model to discern subtle variations in illumination intensity, direction, and color, helping learning better texture-related latent.

In addition to predicting facial attributes, the hierarchical structure of this branch meticulously extracts and preserves geometry features from various layers. These features are coalesced into a latent geometry embedding, denoted as $f_G \in \mathbb{R}^{196 \times 768}$. This embedding plays a pivotal role in guiding the texture estimation process. In the subsequent sections, we will elaborate on the utilization of $f_G$ to enhance the precision and effectiveness of texture synthesis.

\subsection{Geometry-Guided Texture Generator}
\label{sec:VGG-Tex}
As depicted in Fig.~\ref{fig:2 model overview} (bottom), the Geometry-Guided Texture Generator initially employs a vision transformer \cite{ViT} as the backbone, meticulously learning distinct features of local facial regions. In this branch, the input image is segmented into patches and subsequently encoded into latent texture features, designated as the latent texture embedding \( f_T \in \mathbb{R}^{196 \times 768} \). Thereafter, both the latent texture embedding \( f_T \) and the latent geometry embedding \( f_G \) are concurrently fed into the Guidance Attention Block to facilitate the guidance process.

In particular, a cross-attention mechanism is utilized to augment the sensitivity of each latent texture feature to specific attributes within the geometry embedding. This is achieved by computing similarity weight scores through the multiplication of the texture and geometry embeddings. The utilization of this attention mechanism ensures an enhanced alignment between texture and geometry features, effectively mitigating potential discrepancies in facial attributes during rendering.

\begin{equation}
    f_A =  \text{softmax}\left(\frac{(f_T \cdot f_{G}^T)}{\sqrt{d_T}}\right) f_T,
\end{equation}

where \(  f_A  \) represents the attention-enhanced texture embedding matrix. Subsequently, this attention-enhanced texture embedding is processed by the texture decoder to generate the final texture image.

Finally, the texture decoder \( \mathcal{D} \)  integrates the feature \(  f_A  \)  and outputs the texture \( I_T \). 

The overall UV-texture generation process is encapsulated as follows:
\begin{equation}
    I_T = \mathcal{G}(I,f_G),
\end{equation}

where \( I_T \in \mathbb{R}^{1024 \times 1024 \times 3} \) denotes the generated texture image of the input image, \( \mathcal{G} \) is a 2D generative model, which can be either a ViT or a Unet, and \( I \in \mathbb{R}^{256,256,3} \) are the input face images,

\subsection{Visibility-Enhanced Texture Completion Module}

\label{sec:VisibilityEnhancement}

While the dual-branch architecture of VGG-Tex effectively estimates facial textures, it often overlooks critical factors such as occlusions and noise in real-world facial images. These elements can render areas of the image invisible, significantly impacting the quality of the UV texture map reconstruction.

To address these challenges, we propose the Visibility-Enhanced Texture Completion Module. This module leverages a pretrained face-parsing network \cite{faceparsing} to generate a facial skin mask \(M_{\text{skin}}\) for each input image, enhancing the model's capability to manage occlusions effectively.

During the training phase, we implement a selective masking strategy that is intricately guided by the visibility information derived from the facial skin mask associated with each image:
\begin{equation}
M_{\text{mask}} = M_{\text{skin}} \odot B
\end{equation}

where  \( M_{\text{mask}} \) represents the mask applied during training, \( M_{\text{skin}} \) is the facial skin mask obtained from the pretrained face-parsing network, and \( B \) is a random binary mask where specific patches are set to 0 (masked) or 1 (unmasked), based on a predefined probability that controls the density of masking.

This strategy involves the strategic masking of random patches of visible facial skin, creating a targeted learning environment. The purpose of this environment is to intensively prompt the model to concentrate on completing obscured areas, thereby directing its focus towards regions that require specialized attention.

In the testing phase, invisible areas are masked, prompting the model to infer and fill these regions automatically. This phase leverages the learned behaviors from the training phase, where the model has been conditioned to handle and reconstruct occluded or invisible sections of the facial texture.

\subsection{Texture-Guided Geometry Refinement training stage}
\label{sec: TGR}

During our investigations, we have identified that inaccurate landmark fitting, particularly pronounced in scenarios featuring extensive side views, may lead to the overlap of 2D landmarks onto a single pixel. This overlap can detrimentally affect both the quality of geometry reconstruction and texture estimation. To address this challenge, we propose the \textit{Texture-guided Geometry Refinement Module}. This module draws inspiration from the principle of geometry-texture complementarity \cite{oh2001image,blanz2023morphable}, which posits that the interplay between 3D reconstruction's geometry and texture components can be mutually beneficial. According to this principle, not only can the geometry enhance the texture accuracy, but the refined texture can, in turn, further improve the geometric details. 

Specifically, the procedure commences by reconstructing a 3D mesh and UV texture from a given input image \(\mathcal{I}\), subsequently followed by the sampling of a head pose. These poses are quantified by a three-dimensional vector representing rotation angles in the yaw, pitch, and roll dimensions, constrained within the ranges of \([- \frac{\pi}{2}, \frac{\pi}{2}]\), \([- \frac{\pi}{4}, \frac{\pi}{4}]\), and \([- \frac{\pi}{2}, \frac{\pi}{2}]\) respectively. The projection of the 3D head model onto the 2D image space, using the sampled challenging pose, results in the generated input image, designated as \(\mathcal{I}_{r}\). This augmented image is subsequently fed into the geometry prediction module to refine the pose and camera parameters by optimizing the 2D landmarks, thus adapting the model to better accommodate the challenging pose.

In essence, the \textit{Texture-guided Geometry Refinement Module} capitalizes on the synergistic relationship between texture and geometry to enhance the robustness of the model against challenging poses and to improve the accuracy of landmarks. This capability of accommodating diverse poses not only facilitates the generation of refined textures but also aids the model in distinguishing pixel values corresponding to facial features from those representing environmental color elements.

\begin{figure*}
    \centering
    \captionsetup{type=figure}
    \includegraphics[width=\textwidth]{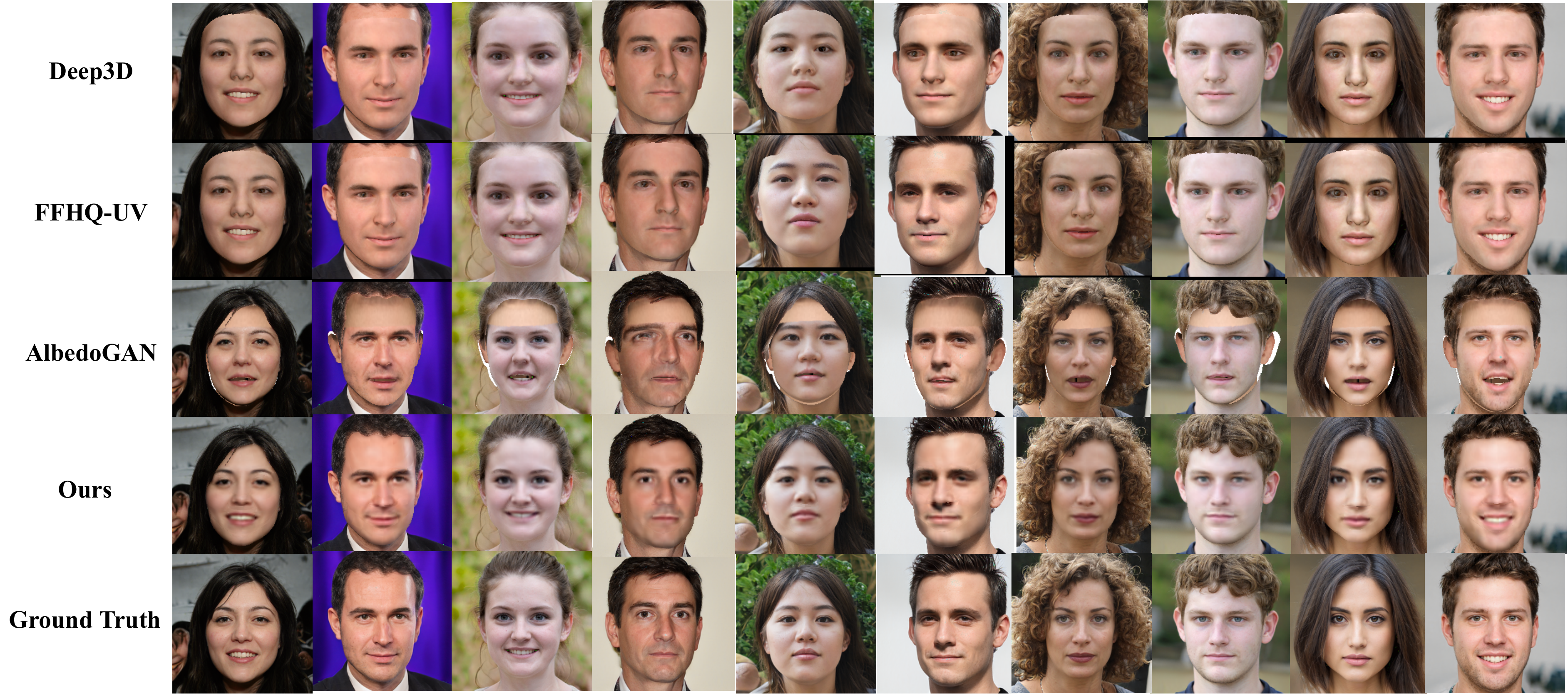}
    \captionof{figure}{\textbf{Comparison of rendering quality to other texture estimation methods.} Our method has the most realistic rendering result and fits into the original image well.}
    \label{quatitative_compare}
    \vspace{-0.2in}
\end{figure*}

\subsection{Loss Function}
Upon the successful acquisition of the 3D geometry face (mesh) and UV texture, these elements are rendered into an output image. Subsequently, we calculate the loss by comparing this output image with the input image, incorporating considerations of the mask and the 2D landmarks. This comparison forms a self-supervised learning cycle, pivotal for the training of our network. In the following sections, we will provide a detailed exposition on the components that constitute our loss function.

\noindent\textbf{Landmark projection Loss.} To optimize shape, expression, and pose parameters, we apply a landmark projection loss. The landmark loss measures the difference between 2D input images and 3D models. The 68 2D landmarks $P_i \in \mathbb{R}^2 (i=1,2,...,68)$ of input images are predicted by PiPNet \cite{pipnet} $\mathbb{M} = \mathbb{R}^{R\times W \times C} \rightarrow \mathbb{R}^{68 \times 2}$, and the corresponding landmarks $M_i \in \mathbb{R}^3 (i=1,2,...,68)$ are selected from the FLAME model surface. Then, selected 3D landmarks are projected onto the 2D space. The landmark loss is defined as
\begin{equation}
L_{lmk}(P,M) = \frac{1}{68} \sum_{i=1}^{68} ||(P_i - \pi(M_i))||_1.
\end{equation}

Besides, we also add L2 regularization to the overall loss to prevent over-fitting
\begin{equation}
L_{reg}(s,e) = ||s||^2_2 + ||e||^2_2
\end{equation}

\noindent\textbf{Rendered Texture Loss.} The rendered texture loss computes the error between the input and rendered images, measuring the difference between ground truth texture and predicted texture. The rendered process can be given as: 
\begin{equation}
    I_{\text{render}} = \mathcal{R}(M, f, T, p),
\end{equation}

where \( \mathcal{R} \) denotes the rendering function, \( M \) is the geometry model, \( f \) is the mapping between UV coordinates and vertex coordinates, \( T \) is the texture image, and \( p \) is the pose parameter.Formally, the loss can be given as:
\begin{equation}
    L_{tex}(I_{\text{input}},I_{\text{render}}) = ||\text{Mask} \odot (I_{\text{input}}-I_{\text{render}})||_{1,1}.
\end{equation}

where $I_{\text{input}}$ is the input image, $I_{\text{render}}$ is the rendered image, $M$ is the mask of the face region adapted from the result of face segmentation method \cite{faceparsing}, setting the face region to 1 and others to 0. $\odot$ denotes the Hadamard product. Taking advantage of differentiable rendering, the loss can be back propagated to the UV texture space.

As mentioned in \ref{sec:VisibilityEnhancement}, visibility-aware texture loss shares the same form as the common texture loss and can be given as:
\begin{equation}
  \begin{aligned}[t]
    &L_{\text{vis\_tex}}(I_{\text{input}}, M, T, p) 
    \\&= \frac{1}{k} \sum_{i=1}^{k} ||\text{Mask}_i \odot (I_{\text{input}} - \mathcal{R}(M, f, T, p_i))||_{1,1}.
  \end{aligned}
\end{equation}

where $k$ denotes the number of different views, $\text{Mask}_i$ and $p_i$ are the mask and pose for diverse views, respectively.

\noindent\textbf{Identity Loss.} To constrain the identity of the predicted texture, we use the features of the face recognition model \cite{arcface} $F: \mathbb{R}^{112 \times 112 \times 3} \rightarrow \mathbb{R}^{512}$. Arcface is trained on 2D images using an additive angular margin loss to obtain highly discriminative features for face recognition. The arcface latent space is invariant to input images' pose, illumination, and other noisy factors. Our identity loss can be defined as the cosine similarity between $I_{\text{input}}$, the input image, and $I_{\text{render}}$, the rendered image:
\begin{equation}
    L_{\text{id}}(I_{\text{input}},I_{\text{render}},F) = \frac{F(I_{\text{input}}) \cdot F(I_{\text{render}})}{||F(I_{\text{input}})||_2\cdot ||F(I_{\text{render}})||_2}.
\end{equation}

\noindent\textbf{Visibility Loss.} During the rendering process, we compute a projection mask $M_{\text{proj}}$ from z-buffer by setting visible pixels to 1, else 0. We optimize the mask error between $M_{\text{skin}}$ and $M_{\text{proj}}$ to minimize the possibility that the texture generator learns from image pixels outside the face region.
\begin{equation}
L_{\text{vis}} = ||M_{\text{proj}}-M_{\text{skin}}||.
\end{equation}

\noindent\textbf{Overall Loss.} The overall loss function $\mathcal{L}$ can be defined as a weighted combination:
\begin{equation}
   \mathcal{L} = L_{\text{lmk}} + L_{\text{tex}} + L_{\text{vis\_tex}} + L_{\text{id}} + L_{\text{reg}} + L_{\text{vis}}.
\end{equation}

\section{Experiments}

\subsection{Implementation Details}

VGG-Tex is trained on the FFHQ \cite{stylegan} and VGGFace2 \cite{vggface2} datasets. The training is conducted on a single RTX 3090 GPU in three phases: first, the Facial Attributes Encoding Module is trained to capture essential facial attributes, followed by joint training with the Geometry-Guided Texture Generator, and finally, the Texture-Guided Geometry Refinement phase. The entire process takes approximately 48 hours with a batch size of 16, using images resized to \(256 \times 256\). Facial regions are extracted using Face Parsing \cite{faceparsing}, and 2D landmarks are detected via PiPNet \cite{pipnet}. The Adam optimizer is used, starting with a learning rate of \(1e^{-3}\), which is reduced by 10\% every 10,000 iterations. In the final phase, the resolution is increased to \(1024 \times 1024\), and an additional 50,000 training steps are performed with a learning rate of \(5e^{-4}\) focusing on refining the Facial Attributes Encoding Module.

\subsection{Comparison on Facial Texture Estimation}

To underscore the superiority of our approach, we commence with a quantitative comparison of our VGG-Tex method against several esteemed benchmarks in the domain of texture synthesis. Specifically, we compare our results with those obtained using DECA \cite{DECA}, TRUST \cite{TRUST}, FFHQ-UV \cite{ffhquv}, Deep3D \cite{deep3d}, and AlbedoGAN \cite{FairAlbedo}. Each of these methods presents innovative designs aimed at enhancing the accuracy of texture estimation. The comparative outcomes are concisely presented in Table \ref{tab:rendering}. The result presented in Table \ref{tab:rendering} distinctly highlights the exceptional performance of our VGG-Tex method in comparison to established benchmarks in the field of texture estimation. Our approach achieves the highest SSIM score of \(0.92 \pm 0.02\), indicating superior structural similarity to the target images, which is crucial for realistic texture synthesis.
\begin{table}[tb]
  \caption{\textbf{Quantitative comparison on texture estimation  on Now Benchmark.} VGG-Tex achieves superior texture estimation performance  to existing strong baselines.}
  \centering
  \begin{tabular}{ccccc}
    \toprule
    & SSIM $\uparrow$ & FID $\downarrow$ & LPIPS $\downarrow$ & ID$\uparrow$ \\
    \midrule
    DECA & 0.30$\pm$0.069 & 81.01 & 0.52$\pm$0.03 & 0.36 \\
    TRUST & 0.30$\pm$0.06 & 111.59 & 0.52$\pm$0.03 & 0.22 \\
    FFHQ-UV & 0.57$\pm$0.28 & 75.70 & 0.33$\pm$0.18& 0.51\\
    Deep3D & 0.84$\pm$0.03 & 67.16 & 0.34$\pm$0.02 & 0.47\\
    AlbedoGAN & 0.82$\pm$0.04 & 67.85 & 0.12$\pm$0.03 & 0.68 \\
    Ours & \textbf{0.92$\pm$0.02 }& \textbf{34.47} & \textbf{0.09$\pm$0.03} & \textbf{0.84} \\
    \bottomrule
  \end{tabular}
  \label{tab:rendering}
 \vspace{-0.2in}
\end{table}

 \begin{table}
   \centering
     \caption{\textbf{Comparison on geometry reconstruction  on NoW benchmark.} VGG-Tex achieves comparable performance  to existing strong baselines. }
   \begin{tabular}{cccc}
     \toprule
     Method  & Median & Mean & Std  \\ \midrule
     Deep3D  & 1.286 & 1.864 & 2.361\\
     DECA    & 1.178 & 1.464 & 1.253\\
    MICA    & \textbf{0.90} & \textbf{1.13} & \textbf{0.95}\\
     Ours    & 0.91  & \textbf{1.13}  & \textbf{0.95}\\
     \bottomrule
   \end{tabular} 
   \label{tab:Now Eval}
   \vspace{-0.1in}
  \end{table}
  
Furthermore, VGG-Tex records the lowest FID score at 34.47, demonstrating that the feature distribution of the generated images closely aligns with that of real images, thereby underscoring the method's effectiveness in producing high-fidelity textures. Additionally, our method outperforms others with a minimal LPIPS score of \(0.09 \pm 0.03\), reflecting a higher perceptual likeness to the original images, an aspect critical for maintaining the visual consistency across different views. Moreover, the Identity Distance (ID) score of 0.84 achieved by VGG-Tex surpasses other methods, affirming its capability in preserving the identity features, which is especially vital in applications involving human faces. These results collectively validate the superiority of VGG-Tex, establishing it as a robust solution for texture estimation that excels across all evaluated metrics, thereby setting a new benchmark in the domain.

Fig. \ref{quatitative_compare} presents a qualitative comparison, showcasing the superior performance of  VGG-Tex  against well-established baselines. It is readily apparent that VGG-Tex not only achieves, but significantly surpasses, the results of competing methods, offering a visually compelling demonstration of its advanced capabilities in facial texture estimation.

\subsection{Comparison on Facial Geometry Reconstruction }
Given our focus on textured 3D face reconstruction, we additionally evaluate the geometry reconstruction quality of our VGG-Tex method by comparing it with established baselines such as Deep3D \cite{deep3d}, DECA \cite{DECA}, and MICA \cite{MICA}. The comparative results are summarized in Table \ref{tab:Now Eval}. It is evident from the results that VGG-Tex achieves performance comparable to the leading model, MICA \cite{MICA}, demonstrating that VGG-Tex not only enhances texture estimation results but also significantly benefits the closely related process of geometry reconstruction. Note that, as mentioned in previous section, the texture of a 3D face can greatly affect how humans perceive it, even if the geometric details are not very fine.

\subsection{Ablations Study}

\begin{table}
  \caption{\textbf{Quantitative ablation study results.}  VTC: Visibility-enhanced Texture Completion module; CG: Geometry-Guidance; LC: Light Condition}
  \label{tab:ablation rendering}
  \centering
  \begin{tabular}{cccc}
    \toprule
    & SSIM $\uparrow$ & FID $\downarrow$ & LPIPS $\downarrow$ \\
    \midrule
    w/o CG & 0.79$\pm$0.04 & 68.82  & 0.14$\pm$0.03 \\
    w/o VTC  & 0.72$\pm$0.04 & 103.24 & 0.20$\pm$0.04 \\
    w/o LC  &  0.81$\pm$0.03 & 43.59  & 0.17$\pm$0.03 \\
    Ours & \textbf{0.92$\pm$0.02 }& \textbf{34.47} & \textbf{0.09$\pm$0.03} \\
    \bottomrule
  \end{tabular}
\end{table}

\begin{figure}
    \centering
    \captionsetup{type=figure}
    \includegraphics[width=0.45\textwidth]{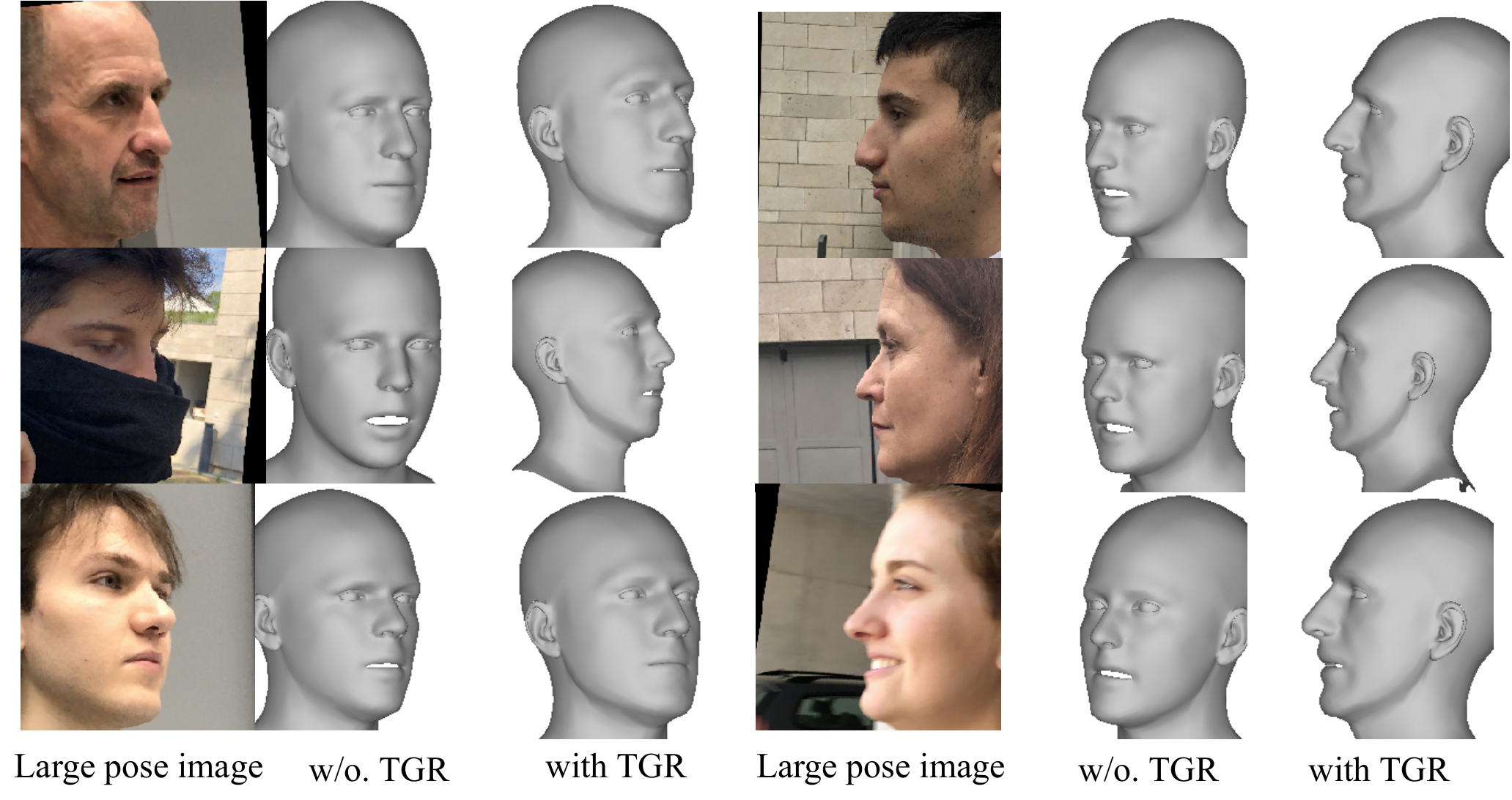}
    \captionof{figure}{\textbf{Qualitative ablation study results.}}
    \label{fig: ablation of texture guidance}
     \vspace{-0.2in}
\end{figure}%

\begin{table}
  \caption{\textbf{Quantitative results of different  Geometry Guided Texture Generator configurations.} Concat: concatenate geometry and texture features. Linear: blend geometry and texture features by MLP. CA: blend geometry and texture features by Cross Attention Module. }
  \label{tab:CGTG results}
  \centering
  \begin{tabular}{cccc}
    \toprule
    CGTG & SSIM $\uparrow$ & FID $\downarrow$ & LPIPS $\downarrow$ \\
    \midrule
    Concat & 0.84 & 50.34 & 0.12 \\
    Linear & 0.87 & 40.21 & 0.10 \\
    CA & \textbf{0.92} & \textbf{34.47} & \textbf{0.09} \\
    \bottomrule
  \end{tabular}
\end{table}
The Facial Attributes Encoding Module m the Geometry Guided Texture Generator, the Visibility-Enhanced Texture Completion module, and the Texture-guided Geometry Refinement training stage are pivotal components of our method. In this section, we explore their efficacy by conducting ablation study.

The Facial Attributes Encoding Module significantly contributes by providing geometric guidance, as evidenced in Table \ref{tab:ablation rendering}. The absence of this guidance notably diminishes performance, confining the texture generation to rely solely on image-derived information. This limitation disregards crucial 3D constraints, thus impacting the precision of texture detail prediction. Additionally, the inclusion of a light condition encoder within this module enhances reconstruction capabilities; its removal, as detailed in the table, similarly leads to a decline in performance. The integration of geometric guidance with texture embedding emerges as a pivotal aspect of the Geometry-Guided Texture Generator. As demonstrated in Table \ref{tab:CGTG results}, substituting cross-attention with alternative operations results in a considerable performance reduction, underscoring the superiority of attention mechanisms. Moreover, the exclusion of the Visibility-Enhanced Texture Completion module, as shown in Table \ref{tab:ablation rendering}, significantly reduces texture estimation efficacy. This is primarily due to its essential role in effectively managing occlusions. To ascertain the advantages of the Texture-guided Geometry Refinement (TGR) training phase, we conduct an ablation study depicted in Fig. \ref{fig: ablation of texture guidance}. The results indicate that models refined through this stage achieve markedly more accurate reconstructions, particularly in scenarios involving extreme head poses.

\section{Conclusion}
This paper introduces VGG-Tex, a novel approach for 3D face reconstruction from monocular images, with a specific emphasis on facial texture estimation. VGG-Tex incorporates several innovative components to enhance performance: the Facial Attributes Encoding Module, the Geometry-Guided Texture Generator, and the Visibility-Enhanced Texture Completion Module. Each of these modules works synergistically to elevate the quality of facial texture estimation. Additionally, the Texture-Guided Geometry Refinement training stage and a novel combined loss function are implemented to optimize the training process. Experimental results have validated the efficacy of our proposed method, demonstrating significant advancements in the field of 3D facial reconstruction.

\newpage
\bibliography{aaai25}

\end{document}


\maketitle

\begin{abstract}
This appendix contains additional materials for the paper “VGG-Tex: A Vivid Geometry-Guided Facial Texture Estimation Model for High  Fidelity Monocular 3D Face Reconstruction”. The appendix is organized as follows: 
\begin{itemize}
  \item Discussion of Texture Representation.
  \item UV Reconstruction and Generation. 
  \item The limitations of VGG-Tex.
  \item Future Work for UV generation.
\end{itemize}
\end{abstract}

\section{Discussion of Texture Representation}
3D head reconstruction has gained impressive progress in recent years. Current methods can be Recent years have witnessed significant progress in 3D head reconstruction, with existing methods generally categorized into explicit and implicit representations. 
Explicit methods, which define graphical representations for head textures, are typically more interpretable and resource-efficient. A dominant approach in this category is UV mapping, where texture information is encoded by storing the RGB details of a 3D mesh model in a 2D image. These color values are then re-projected onto the vertices of the mesh.

Recently, 3D Gaussian Splatting (3DGS)\cite{qian20243dgs,xu2024gaussian,qian2024gaussianavatars,} have gained increasing attention. Unlike traditional approaches, 3DGS represents a 3D model as a set of 3D Gaussians, which leads to a fundamentally different pipeline for rasterization and rendering.

In contrast, implicit methods, particularly those based on Neural Radiance Fields (NeRF)\cite{mildenhall2021nerf,hong2022headnerf,xu2023avatarmav,ma2023otavatar}, implicitly model both color density and spatial locations, enabling the production of highly realistic 3D reconstructions.
These methods each have distinct advantages and disadvantages.

\textbf{UV-based Representations:} UV-based methods are \textbf{intuitive} and compatible with traditional 2D image processing techniques, allowing them to represent facial details with high fidelity, similar to 2D images. However, their effectiveness is heavily reliant on the quality of the UV mapping process and the underlying 3D mesh model. The representational capability is mainly constrained by the number of vertices in the 3D mesh, which can limit the detail and accuracy of the reconstruction. Despite these limitations, UV maps remain widely used due to their interpretability and ease of integration with existing graphics pipelines.

\textbf{3DGS-based Representations:} The 3D Gaussian Splatting (3DGS) approach represents both geometry and color information simultaneously through a set of 3D Gaussians.  One of the key advantages of 3DGS is its ability to handle complex scenes with varying levels of detail and resolution, providing a flexible and efficient representation that is well-suited for real-time rendering applications.

\textbf{NeRF-based Representations:} NeRF-based methods employ neural networks to implicitly represent 3D information and render the 3D model using ray tracing techniques. These methods excel at producing photorealistic results by modeling both the color and density of points in 3D space. However, they often require extensive computational resources and long training times, making them less practical for real-time applications. Additionally, NeRF-based models can struggle with fine details and may require a large amount of data to achieve high-quality reconstructions.

The reason we chose UV representation for our head texture generation is to take advantage of existing 3DMM models and 3D geometry reconstruction results. At the same time, try to exhaust the representation ability in image space.

\section{UV Reconstruction and Generation}
In this paper, we introduce a novel method for 3D head reconstruction with the primary objective of faithfully capturing the texture details of an individual. Our approach is designed as a deterministic process, in contrast to generative tasks that focus on modeling the data distribution of UV images. Consequently, GAN-based and diffusion-based models are not well-suited for accurately reconstructing human head textures. However, UV generation remains a valuable area of research. Accurate reconstruction can provide high-quality data that could significantly enhance generative tasks, which we leave as a direction for future work.

\section{The Limitations of VGG-Tex}
Although VGG-Tex can reconstruct detailed UV textures for 3D mesh models, our method is still constrained by the limitations of monocular face image datasets. As a result, there are quality discrepancies between visible and invisible areas on the 3D model. Furthermore, our approach is designed to learn a deterministic mapping from the face domain to the UV domain, which inherently limits its ability to perform unconditional generation. This limitation may impact its applicability in certain downstream tasks.

\section{Future Work}
There are two key aspects to improving texture quality in head generation. First, adapting 3D Gaussian Splatting (3DGS) and other forms of representation can address the limitations inherent in 3D mesh models. Second, combining generative models with related theories can extend reconstruction models to simultaneously support both reconstruction and generation tasks.

\section{Reproducibility Checklist}
This paper:
Includes a conceptual outline and/or pseudocode description of AI methods introduced (yes/partial/no/NA) yes

Clearly delineates statements that are opinions, hypothesis, and speculation from objective facts and results (yes/no) yes

Provides well marked pedagogical references for less-familiare readers to gain background necessary to replicate the paper (yes/no) yes

Does this paper make theoretical contributions? (yes/no) no

If yes, please complete the list below.

All assumptions and restrictions are stated clearly and formally. (yes/partial/no)

All novel claims are stated formally (e.g., in theorem statements). (yes/partial/no)

Proofs of all novel claims are included. (yes/partial/no)

Proof sketches or intuitions are given for complex and/or novel results. (yes/partial/no)

Appropriate citations to theoretical tools used are given. (yes/partial/no)

All theoretical claims are demonstrated empirically to hold. (yes/partial/no/NA)

All experimental code used to eliminate or disprove claims is included. (yes/no/NA)

Does this paper rely on one or more datasets? (yes/no) yes

If yes, please complete the list below.

A motivation is given for why the experiments are conducted on the selected datasets (yes/partial/no/NA) partial.

All novel datasets introduced in this paper are included in a data appendix. (yes/partial/no/NA) yes.

All novel datasets introduced in this paper will be made publicly available upon publication of the paper with a license that allows free usage for research purposes. (yes/partial/no/NA) yes.

All datasets drawn from the existing literature (potentially including authors’ own previously published work) are accompanied by appropriate citations. (yes/no/NA) yes. 

All datasets drawn from the existing literature (potentially including authors’ own previously published work) are publicly available. (yes/partial/no/NA) yes. 

All datasets that are not publicly available are described in detail, with explanation why publicly available alternatives are not scientifically satisficing. (yes/partial/no/NA) NA.

Does this paper include computational experiments? (yes/no) yes.

If yes, please complete the list below.

Any code required for pre-processing data is included in the appendix. (yes/partial/no). yes.

All source code required for conducting and analyzing the experiments is included in a code appendix. (yes/partial/no) paritial. 

All source code required for conducting and analyzing the experiments will be made publicly available upon publication of the paper with a license that allows free usage for research purposes. (yes/partial/no) paritial.

All source code implementing new methods have comments detailing the implementation, with references to the paper where each step comes from (yes/partial/no)  yes.

If an algorithm depends on randomness, then the method used for setting seeds is described in a way sufficient to allow replication of results. (yes/partial/no/NA)  yes.

This paper specifies the computing infrastructure used for running experiments (hardware and software), including GPU/CPU models; amount of memory; operating system; names and versions of relevant software libraries and frameworks. (yes/partial/no) yes. 

This paper formally describes evaluation metrics used and explains the motivation for choosing these metrics. (yes/partial/no) yes.

This paper states the number of algorithm runs used to compute each reported result. (yes/no) yes.

Analysis of experiments goes beyond single-dimensional summaries of performance (e.g., average; median) to include measures of variation, confidence, or other distributional information. (yes/no) yes.

The significance of any improvement or decrease in performance is judged using appropriate statistical tests (e.g., Wilcoxon signed-rank). (yes/partial/no) no.

This paper lists all final (hyper-)parameters used for each model/algorithm in the paper’s experiments. (yes/partial/no/NA) yes.

This paper states the number and range of values tried per (hyper-) parameter during development of the paper, along with the criterion used for selecting the final parameter setting.   (yes/partial/no/NA) partial. 

\bibliography{aaai25}